\documentclass[final]{cvpr}

\usepackage{times}
\usepackage{epsfig}
\usepackage{graphicx}
\usepackage{amsmath}
\usepackage{amssymb}

\newif\ifdraft

\newcommand{\bt}{\mathbf{t}}
\newcommand{\bM}{\mathbf{M}}
\newcommand{\br}{\mathbf{r}}

\newcommand{\bX}{\mathbf{X}}
\newcommand{\bz}{\mathbf{z}}

\newcommand{\bB}{\mathbf{B}}
\newcommand{\bS}{\mathbf{S}}
\newcommand{\bH}{\mathbf{H}}
\newcommand{\bR}{\mathbf{R}}

\newcommand{\bdelta}{\boldsymbol{\delta}}

\newcommand{\ms}[1]{\ifdraft {\color{green}{#1}} \else {#1}\fi}
\newcommand{\kn}[1]{\ifdraft {\color{blue}{#1}} \else {#1}\fi}

\newcommand{\MS}[1]{\ifdraft {\color{green}{\textbf{MS: #1}}}\else {}\fi}
\newcommand{\KN}[1]{\ifdraft {\color{red}{\textbf{KN: #1}}}\else {}\fi}

\newcommand{\comment}[1]{}

\iftrue 

\else 

\fi

\usepackage{booktabs}
\usepackage{multirow}
\usepackage{bbding}
\usepackage{pifont}

\usepackage[pagebackref=true,breaklinks=true,colorlinks,bookmarks=false]{hyperref}

\def\cvprPaperID{XX} 
\def\confYear{CVPR 2021}

\begin{document}

\title{Temporally-Transferable Perturbations: Efficient, One-Shot Adversarial Attacks for Online Visual Object Trackers}

\author{%
	\vspace{0.5em}
	{Krishna Kanth Nakka $^1$ and Mathieu Salzmann $^{1,2}$} \\
	{ $^1$ EPFL Computer Vision Lab, \quad $^2$ ClearSpace SA} \\
}

\maketitle


\begin{abstract}
	
	In recent years, the trackers based on Siamese networks have emerged as highly effective and efficient for visual object tracking (VOT). While these methods were shown to be vulnerable to adversarial attacks, as most deep networks for visual recognition tasks, 
	the existing attacks for VOT trackers all require perturbing the search region of every input frame to be effective, which comes at a non-negligible cost, considering that VOT is a real-time task.
	In this paper, we propose a framework to generate a single temporally-transferable adversarial perturbation from the object template image only. This perturbation can then be added to every search image, which comes at virtually no cost, and still successfully fool the tracker. Our experiments evidence that our approach outperforms the state-of-the-art attacks on the standard VOT benchmarks in the untargeted scenario. Furthermore, we show that our formalism naturally extends to targeted attacks that force the tracker to follow any given trajectory by precomputing diverse directional perturbations.
	

\end{abstract}


\section{Introduction}

Visual Object Tracking (VOT)~\cite{vot2015} is a key component of many vision-based systems, such as surveillance and autonomous driving ones. Studying the robustness of object trackers is therefore critical from a safety and security point of view. When using deep learning, as most modern tackers do, one particular security criterion is the robustness of the deep network to adversarial attacks, that is, small perturbations aiming to fool the prediction of the model. In recent years, the study of such adversarial attacks has become an increasingly popular topic, extending from image classification~\cite{goodfellow2014explaining,kurakin2016adversarial} to more challenging tasks, such as object detection~\cite{xie2017adversarial} and  segmentation~\cite{arnab2018robustness,fischer2017adversarial}.


\begin{figure}[t]
	\footnotesize
	\centering
      \includegraphics[width=8cm]{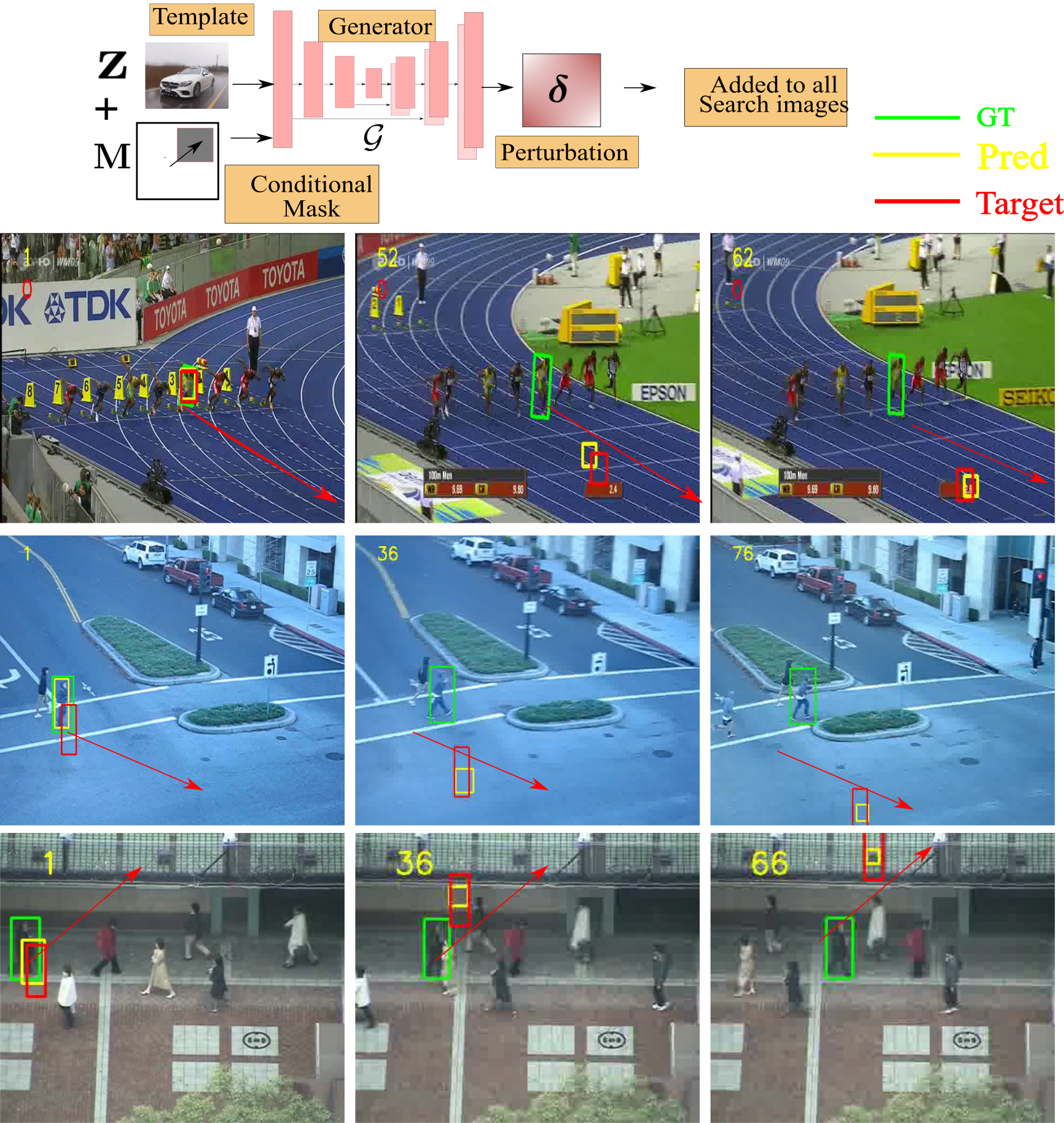}
	\caption{{Our approach generates a single transferable perturbation from the object template in \textbf{8}ms. We can then use this perturbation to attack the tracker throughout the {\it entire} sequence, either in an untargeted manner, or a targeted one by forcing it to follow a predefined motion, such as a fixed direction as illustrated above, or a more complicated trajectory, as shown in our experiments.}}
	\label{fig:teaser}
\end{figure}

VOT is no exception to this rule. However, unlike in recognition-driven tasks, where the predictions are typically restricted to a predefined label set, such as image categories in ImageNet or object categories in PASCAL VOC, the category of the objects tracked at test time by VOT methods is not known beforehand and only defined by the first frame of the input video. In other words, adversarial attacks on VOT systems should generalize to object templates that were not seen during the training phase.  Several methods~\cite{spark,oneshot2020,jia2019fooling,csa,fan} have proposed solutions to address this challenging scenario, designing attacks to fool the popular Siamese-based trackers~\cite{siamrpnpp,siamfc,siamrpn,siammask}. Among these, however, the attacks in~\cite{spark,oneshot2020,jia2019fooling} are either too time-consuming or designed to work on the entire video, thus not applicable to fooling a tracker in real time and in an online fashion. By contrast,~\cite{csa,fan} leverage generative methods~\cite{poursaeed2018generative,xiao2018generating} to synthesize perturbations in real time and thus can effectively attack online trackers. Nevertheless, these methods still require generating a perturbation for the search region of every individual input frame, {which causes an undesirable computational burden.}
{This burden becomes particularly pronounced for long-term trackers, because the search region grows to incorporate larger context in the presence of occlusions or when losing track of the object, leading to an increased attack time.}

In this paper, we address this by introducing a temporally-transferable attack strategy. Specifically, we learn to generate a single, one-shot perturbation that is transferable across all the frames of the input video sequence. To this end, unlike existing works~\cite{csa,fan} that generate perturbations for the template and for every search image independently, we train a perturbation generator to synthesize a \emph{single} perturbation from the template, and add this perturbation to every subsequent search image. Because the perturbation is generated only once from the template, applying it to the larger search regions of long-term trackers comes at no additional cost.

Furthermore, in contrast to existing frameworks~\cite{csa,fan}, our approach naturally extends to performing \emph{targeted} attacks so as to steer the tracker to follow any specified trajectory.  To this end, we condition our perturbation generator on the targeted direction and train the resulting conditional generator to produce perturbations that correspond to arbitrary, diverse input directions. At test time, we can then pre-compute perturbations for a small number of diverse directions, \comment{e.g., 8 directions at $45^{\circ}$ ,}
e.g., 12 in our experiments,
and apply them in turn so as to generate the desired complex trajectory.
{We illustrate this in Fig.~\ref{fig:teaser}, where a single perturbation computed in 8 ms can steer the tracker to move along a given direction for the entire video sequence, and will show more complex trajectories in our experiments.}

Overall, our contributions can be summarized as follows:
\begin{itemize}
	\vspace{-0.2cm}
	\item We introduce a temporally-transferable attack strategy to fool Siamese-based trackers by generating a single, one-shot perturbation from the template only.
	\vspace{-0.2cm}
	\item Our approach is applicable in an online manner, requiring only 8ms to generate our one-shot perturbation, thus maintaining real-time performance. 
	\vspace{-0.2cm}
	\item Our framework naturally extends to targeted attacks, and our conditioning scheme allows us to steer the tracker to follow complex, erroneous trajectories. In practice, this would let one generate plausible incorrect tracks, making it harder to detect the attack.
\end{itemize}

We demonstrate the benefits of our approach on 4 benchmark datasets, namely, OTB100~\cite{otb100}, VOT2018~\cite{vot2018} and UAV123~\cite{uav123} and  VOT2018-LT~\cite{vot2018}. Our experiments show that our approach yields more effective attacks than the state-of-the-art method~\cite{csa}, even though it needs to attack every individual video frame while we don't. We will make our code publicly available.


\section{Related Work}

\noindent{\textbf{Visual Object Tracking.}} VOT aims to estimate the position of a template cropped from the first frame of a video in each of the subsequent video frames. Unlike most other visual recognition tasks, such as image classification or object detection, that rely on predefined categories, VOT seeks to generalize to any target object at inference time. As such, early works mainly focused on measuring the correlation between the template and the search image~\cite{cf1}, extended to exploiting multi-channel information~\cite{kiani2013multi} and spatial constraints~\cite{kiani2015correlation,danelljan2015learning}. 

Nowadays, VOT is commonly addressed by end-to-end learning strategies. In particular, Siamese network-based trackers~\cite{siamfc,siamrpn,siammask,siamrpnpp,DAsiamrpn} have grown in popularity because of their good speed-accuracy tradeoff and generalization abilities. 
The progress in this field includes the design of a cross-correlation layer to compare template and search image features~\cite{siamfc}, the use of a region proposal network (RPN)~\cite{ren2016faster} to reduce the number of multi-scale correlation operations~\cite{siamfc}, the introduction of an effective sampling strategy to account for the training data imbalance~\cite{DAsiamrpn}, the use of multi-level feature aggregation and of a spatially-aware sample strategy to better exploit deeper ResNet backbones~\cite{li2019siamrpnpp}, and the incorporation of an auxiliary segmentation training objective to improve the tracking accuracy~\cite{siammask}. In our experiments, we will focus on the SiamRPN++~\cite{siamrpnpp} tracker due to its popularity in real-world applications. Nevertheless, we will study the transferability of our generated adversarial attacks to other trackers.

\noindent{\textbf{Adversarial Attacks.}} Adversarial attacks were first investigated in~\cite{szegedy2013intriguing} to identify the vulnerability of modern deep networks to  imperceptible perturbations in the context of image classification. Different attack strategies were then studied, including single step gradient descent~\cite{goodfellow2014explaining,kurakin2016adversarial}, DeepFool~\cite{moosavi2016deepfool,dong2018boosting}, and computationally more expensive attacks, such as CW~\cite{carlini2017towards}, JSMA~\cite{jsma}, and others~\cite{croce2019sparse,su2019one,narodytska2017simple}. As of today, PGD~\cite{madry2017towards} is considered as one of the most effective attack strategies, and has been employed to other visual recognition tasks, such as object detection~\cite{xie2017adversarial} and semantic segmentation~\cite{arnab2018robustness}.

Inspired by this progress, iterative adversarial attacks have been studied in the context of VOT. In particular, SPARK~\cite{guo2019spark} computes incremental perturbations by using information from the past frames;~\cite{chen2020one} exploits the full video sequence to attack the template by solving an optimization problem relying on a dual attention loss. While effective, most of the above-mentioned attacks are time-consuming, because of their use of heavy gradient computations or iterative schemes. As such, they are ill-suited to attack an online visual tracking system in real time. {\cite{PAT} also relies on a gradient-based scheme to generate a physical poster that will fool a tracker. While the attack is then real-time, it requires the attacker to physically alter the environment.}

As an efficient alternative to iterative attacks,  AdvGAN~\cite{advgan} proposed to train a generator that synthesizes perturbations in a single forward pass. Such generative perturbations were extended to VOT in~\cite{csa,fan}. For these perturbations to be effective, however, both~\cite{csa} and~\cite{fan} proposed to attack every individual search image, by passing it through the generator. To be precise, while~\cite{csa} studied the problem of attacking the template only the success of the resulting attacks was shown to be significantly lower than that of perturbing each search image. Doing so, however, becomes increasingly expensive as the search region grows, as can happen in long-term tackers. Here, we show that temporally-transferable perturbations can be generated in a one-shot manner from the template, yet still effectively fool the VOT system in every search image; the success of our approach lies in the fact that, while the perturbation is generated from the template, it is added to every search image. This addition, however, comes at virtually no cost, as it does not involve inference in a network. Furthermore, our approach can be extended to producing targeted attacks by conditioning the generator on desired directions. In contrast to~\cite{fan}, which only briefly studied targeted attacks in the restricted scenario of one specific pre-defined trajectory, our approach allows us to create arbitrary, complex trajectories at test time, by parametrizing them in terms of successive targeted perturbations.


\section{Method}

In this section, we introduce our temporally-transferable attack framework for object trackers based on Siamese networks. The goal of our method is to efficiently generate a single perturbation from the object template, which we can then transfer to adversarially attack the subsequent video frames. Below, we formalize the task of performing adversarial attacks on an object tracker, and then introduce our one-shot perturbation strategy in detail, first for untargeted attacks and then for targeted ones.

\subsection{Problem Definition}

Let  $\bX$ = $\{\bX_i\}_1^T$ denote the frames of a video sequence of length $T$, and $\bz$ be the template cropped from the first frame of the video.  A tracker $\mathcal{F}(\cdot)$ aims to locate the template $\bz$ in search regions extracted from the subsequent video frames.  Since the tracked target generally does not move wildly between two consecutive frames, most trackers only search for the target in a small region  $\bS_i$ centered at the position obtained in the previous frame. At each time-step, the template $\bz$ and the search region $\bS_i$ are first passed individually through a shared backbone network, and the resulting features are processed by some non-shared  layers and fused by depth-wise separable correlation.  
The fused features then act as input to a region proposal network (RPN), which predicts a classification map $\bH\in\mathbb{R}^{H \times W \times K}$ and a bounding box regression map $\bR\in\mathbb{R}^{H \times W \times 4K}$, where $K$ denotes the bounding box anchors at each spatial location. In short, at each spatial location, the classification map encodes the probability of each anchor to contain the target, and the regression map produces a corresponding offset for the anchor location {and size}. In practice, most trackers further re-rank the proposals be exploiting both a Gaussian window, which penalizes the proposals that are away from the region center, and scale change penalty, which prevents the object size from varying too strongly from that in the previous frame.

Formally, given the template $\bz$, we  aim to find a temporally-transferable perturbation $\boldsymbol{\delta}$  that, when added to any search region $\bS_i$ to obtain an adversarial image $\tilde{\bS}_i = \bS_i + \bdelta$, leads to an incorrect target localization in frame $i$.
In the untargeted attack scenario, we aim for the bounding box to simply deviate from the trajectory predicted by the unattacked tracker, thereby losing the tracked object. In the case of a targeted attack, we seek to force the center of the tracked window to follow a trajectory specified by the attacker, which can be either a simple fixed, yet arbitrary direction, or a more complicated trajectory.


\begin{figure}[t]
	\footnotesize
	\centering
      \includegraphics[width=\linewidth]{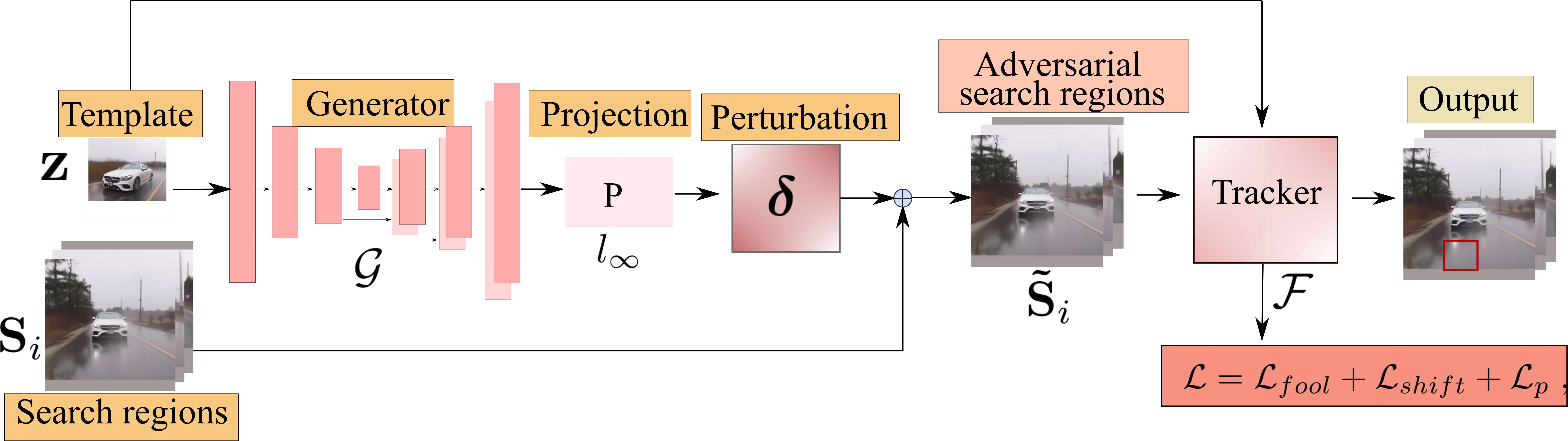}
	\caption{{{\bf Our temporally-transferable attack framework.} We generate a single perturbation from the template, and add it to the search region of every subsequent frame so as to fool the tracker.}} 
	\label{fig:frameworkfull}
\end{figure}

\subsection{One-Shot Perturbation Generator}
\noindent{\textbf{Pipeline.}} Figure~\ref{fig:frameworkfull} illustrates the overall architecture of our framework, which consists of two main modules: a generator $\mathcal{G}$ and tracker $\mathcal{F}$. 
Unlike existing works~\cite{csa,fan},
we train our perturbation generator to synthesize a \emph{single} perturbation  $\bdelta$  from the template, and simply add this perturbation to every subsequent search image. To train the generator, we therefore extract a template $\bz$ from the first frame of a given video sequence and feed it to the generator to obtain a perturbation $\bdelta$. We then crop $N$ search regions from the subsequent video frames 

using ground-truth information, and add $\bdelta$ to each such regions to obtain adversarial search regions $\tilde{\bS} =  \{\tilde{\bS}_i\}_1^N$. 
Finally, we feed the clean template $\bz$ and each adversarial search region $\tilde{\bS}_i$ to the tracker to produces an adversarial classification map $\tilde{\bH}_i\in\mathbb{R}^{H \times W \times K}$ \KN{changed from 2K to K} and regression map $\tilde{\bR}_i\in\mathbb{R}^{H \times W \times 4K}$.  

Our goal is for these adversarial classification and regression maps to fool the tracker, i.e., result in erroneously locating the target. To this end, we compute the classification map $\bH_i$ for the unperturbed search image, and seek to decrease the score in $\tilde{\bH}_i$ of any proposal $j$ such that $\bH_i(j) > \tau$, where $\bH_i(j)$ indicates the probability for anchor $j$ to correspond to the target\footnote{With this notation, we assume, w.l.o.g., that $j$ represents a specific anchor at a specific spatial location.} and $\tau$ is a threshold. 
Following~\cite{csa}, we achieve this by training the perturbation generator $\mathcal{G}$ with the adversarial loss term 
\begin{equation}\label{eq:foolingloss}
\begin{aligned}
\small
\mathcal{L}_{fool}(\mathcal{F}, \bz, \tilde{\bS}_i) =  \lambda_1 \hspace{-0.2cm}  \sum_{j | \bH_i(j)>\tau} \hspace{-0.2cm} \max\left( \tilde{\bH}_i(j)- (1-\tilde{\bH}_{i}(j)) , \mu_c\right) \\
+ \lambda_2 \hspace{-0.2cm}	 \sum_{j | \bH_i(j) >\tau} \hspace{-0.2cm} \left( \max\left( \tilde{\bR}_i^{w}(j) , \mu_w\right) + \max\left( \tilde{\bR}_i^{h}(j), \mu_h\right)\right)\;,
\end{aligned}
\end{equation}

where  

$\tilde{\bR}_i^{w}(j)$ and $\tilde{\bR}_i^{h}(j)$ represent the width and height regression values for anchor $j$.  The first term in this objective aims to simultaneously decrease the target probability and increase the background probability for anchor $j$ where the unattacked classification map contained a high target score. The margin $\mu_c$ then improves the numerical stability of this dual goal. The second term encourages the target bounding box to shrink, down to the limits $\mu_w$ and $\mu_h$, to facilitate deviating the tracker. 

\subsubsection{Untargeted Attacks}

The loss $\mathcal{L}_{fool}$ discussed above only aims to decrease the probability of the anchors obtained from the unattacked search region. Here, we propose to complement this loss with an additional objective seeking to explicitly activate a different anchor box, which we will show in our experiments to improve the attack effectiveness.
Specifically, we aim for this additional loss to activate an anchor away from the search region center, so as to push the target outside the true search region, which ultimately will make the tracker be entirely lost. To achieve this, we seek to activate {an
anchor} $t$ lying at a distance $d$ from the search region center. We then write the loss

\begin{equation}\label{eq:shiftloss}
\begin{aligned}
\small
\mathcal{L}_{shift}(\mathcal{F}, \bz, \tilde{\bS}_i) =  \lambda_3   L_{cls}(\tilde{\bH}_i(t)) + \lambda_4 L_{reg}(\tilde{\bR}_i(\bt), \br^*) \;,
\end{aligned}
\end{equation}
where  $L_{cls}$ is a classification loss encoding the negative log-likelihood of predicting the target at location $t$, and $L_{reg}$ computes the $L_1$ loss between the regression values at location $t$ and pre-defined regression values $\br^*\in \mathbb{R}^4$.

\subsubsection{Targeted Attacks}\label{sec:targeted}
The untargeted shift loss discussed above aims to deviate the tracker from its original trajectory. However, it does not allow the attacker to force the tracker to follow a pre-defined trajectory. To achieve this, we modify our perturbation generator to be conditioned on the desired direction we would like the tracker to predict. In practice, we input this information to the generator as an additional channel, concatenated to the template.  

Specifically,  we  compute  a binary mask  $\bM_i \in \{0,1\} ^ {(W \times H )}$,  and set $\bM_i(j)=1$ at all spatial locations under the bounding box which we aim the tracker to output. Let $\bB_i^{t}$ be such a targeted bounding box, and $\br_i^t$ the corresponding desired offset from the nearest anchor box. 

We can then express a shift loss similar to the one in Eq.~\ref{eq:shiftloss} but for the targeted scenario as
\begin{equation}\label{eq:targetloss}
\begin{aligned}
\small
\mathcal{L}_{shift}(\mathcal{F}, \bz, \tilde{\bS}_i, \bM_i) =  \lambda_3   L_{cls}(\tilde{\bH}_i(t)) + \lambda_4 L_{reg}(\tilde{\bR}_i(t), \br_i^t)\;, \\
\end{aligned}
\end{equation}
where, with a slight abuse of notation, $t$ now encodes the targeted anchor.

\subsubsection{Overall Loss Function}

In addition to the loss functions discussed above, we use a perceptibility loss $\mathcal{L}_p$  aiming to make the generated perturbations invisible to the naked eye. We express this loss as
\begin{equation}
\mathcal{L}_{p} = \lambda_5  \| \mathcal{\bS}_i- \textrm{Clip}_{\{\bX_i, \epsilon\}}\{\mathcal{\bS}_i+ \bdelta\} \|_{2}^2\;,\label{L2loss}
\end{equation}
where the $\textrm{Clip}$ function enforces an $L_\infty$ bound $\epsilon$ on the perturbation.

We then write the complete objective to train the generator as
\begin{equation}
\begin{aligned}
\mathcal{L}(\mathcal{F},\bz, \bS_i) =   \mathcal{L}_{fool}+ \mathcal{L}_{shift} + \mathcal{L}_{p}\;,
\end{aligned}
\label{eq:eq9}
\end{equation}
where $\mathcal{L}_{shift}$ corresponds to Eq.~\ref{eq:shiftloss} in the untargeted case, and to Eq.~\ref{eq:targetloss} in the targeted one.

\subsection{Attacking the Tracker at Inference Time} 
Once the generator is trained using the loss in Eq.~\ref{eq:eq9}, we can use it to generate a temporally-transferrable perturbation from the template of any new test sequence, and use the resulting perturbation in an online-tracking phase. Generating such a perturbation takes only $8$ ms, and adding it to the subsequent frames is virtually free.

For targeted attacks, in the simple case where the desired trajectory is a constant direction, we condition the generator on a specific mask $\bM$, which activates  a square patch centered along the desired direction and at a given offset from the center. To force the tracker to follow more complex trajectories during inference, such as following the ground-truth trajectory with an offset, we precompute perturbations for a small number, $K$, of predefined, diverse directions, with 
\comment{8 directions at 45$^\circ$ of each other}
$K=12$ in our experiments.

 We then define the desired complex trajectory as succession of these predefined directions.


\section{Experiments}

\noindent{\textbf{Datasets.}} Following~\cite{csa}, we train our perturbation generator on GOT-10K~\cite{got10k} and evaluate its effectiveness on 3 short-term tracking datasets, OTB100~\cite{otb100},  VOT2018~\cite{vot2018} and UAV123~\cite{uav123}, and on one long-term benchmark VOT2018-LT~\cite{vot2018}.  We primarily use the SiamRPN~\cite{siamrpn} and SiamRPN++~\cite{siamrpnpp} trackers with AlexNet~\cite{alexnet}  
and~\cite{resnet} backbones, respectively, and train our generator to perform white-box attacks. We also study our our attacks transfer to other popular trackers in Section~\ref{sec:transferability}.

\noindent{\textbf{Evaluation Metrics.}} As in~\cite{csa,spark,chen2020one,fan}, we report the performance of our adversarial attacks using the metrics employed by each dataset to evaluate the effectiveness of unattacked trackers.

Specifically, for OTB100~\cite{otb100} and UAV123~\cite{uav123}, we report the  precision (P) and success  scores (S). The precision encodes the proportion of frames for which  the center of the tracking window is within 20 pixels of the {ground-truth center}. 
The success corresponds to the proportion of frames for which the overlap between the predicted and ground-truth {tracking window}
is greater than a given threshold.
For VOT2018, we report the Expected Average Overlap (EAO), a measure that considers both the accuracy (A) and robustness  (R) of a tracker. Specifically, the accuracy denotes the average overlap, and the robustness is computed from the number of tracking failures. Furthermore, we also report the number of restarts because the standard VOT evaluation protocol reinitializes the tracker once it is too far away from the ground truth.  For VOT2018-LT~\cite{vot2018}, we report the tracker precision,  recall, and  F1-score.  To evaluate targeted attacks, we further report the {proportion of frames in which the predicted center and the target trajectory's center are at a distance of at most 20 pixels for short-term datasets and 50 pixels for VOT2018-LT~\cite{vot2018}.}

\noindent{\textbf{Implementation Details.}}  We implement our approach in PyTorch~\cite{pytorch} and perform our experiments on an NVIDIA Telsa V100 GPU with 32GB RAM.  We  train the generator using pre-cropped search images uniformly sampled every 10 frames from the video sequences of GOT-10K~\cite{got10k}. We use the Adam~\cite{adam} optimizer with a learning rate of  $2 \times 10^{-4}$. We set the margin thresholds $m_c$, $m_w$, $m_h$ to -5 as in~\cite{csa}, and the $l_{\infty}$ bound  $\epsilon$  to \{8, 16\}.  To fool the tracker, we use $\lambda_1=0.1$, $\lambda_2=1$ in Eq~\ref{eq:foolingloss}, and activate an anchor at distance  $d=4$ directly below the true center and of size $64\times64$ for all untargeted experiments. \kn{ We set $\lambda_5$ to 500 for all experiments.} \MS{There is an inconsistency here. First $\lambda_5=500$, then $\lambda_5 = 1$.}\KN{corrected  the lambdas}
Furthermore, we set the shift loss weights $\lambda_3$ and $\lambda_4$ to 0.1 and 1, respectively. For targeted attacks, we define $\br_i^t$ in Eq.~\ref{eq:targetloss} as a randomly-selected anchor at distance $d=4$ from the true center and set its size to $64\times64$ for all datasets, except for VOT2018-LT where we use $90\times90$.  We resize the search images to $255 \times 255$  and the template to $127 \times 127$ before passing them to the tracker. For SiamRPN++\cite{siamrpnpp}, the final feature map will be of size $W=25$, $H=25$. 

\noindent{\textbf{Methods.}} We compare our approach with the state-of-the-art, {generator-based} CSA~\cite{csa} attack strategy, {which also performs online attacks.} CSA can be employed in 3 settings:  CSA\_T, which on attacks the template, CSA\_S, which attacks all search images, and  CSA\_TS, which attacks both the template and all search regions.  
We denote our temporally-transferable attacks obtained from the total loss in Eq.~\ref{eq:eq9} as $\textrm{Ours}$ and a variant of our method without the term $\mathcal{L}_{shift}$ but only with $\mathcal{L}_{fool}$ as $\textrm{Ours}_f$.

Unlike CSA~\cite{csa}, which takes 3ms to attack the template and  8ms to compute a perturbation for each search region, our approach generates a single transferable perturbation from the template in 8ms, and uses it at virtually no additional cost\footnote{The additional cost comes from adding the perturbation to the search images, an operation that CSA\_S and CSA\_TS must also perform.} for the \emph{entire} video sequence to attack in an untargeted manner.  For targeted attacks, we precompute $K=12$ diverse directional perturbations, each taking 8ms to generate, and use them to force the tracker to follow any target trajectory. In our experiments, we also report the average attack cost per video over the full dataset 
for all approaches.

\subsection{Untargeted Attacks}\label{sec:untargeted attacks}

\noindent{\bf Results on OTB100.}  In Table~\ref{table:untargeted_otb}, we compare the results of our approach and of the different settings of CSA for both the SiamRPN and SiamRPN++ trackers. Our method {works at a speed comparable to} CSA\_T
but significantly outperforms it in all cases.  Furthermore, thanks to our shift loss, our attack becomes even stronger than CSA\_S and CSA\_TS, while requiring only a fraction of their computation budget. For example, on SiamRPN++~\cite{siamrpnpp}, $\textrm{Ours}$ reduces the precision of the tracker to $8\%$, compared to  $51\%$ for CSA\_TS.

\begin{table}[t]
	\centering
	\resizebox{0.96\columnwidth}{!}{
		\begin{tabular}{cccccccccc}
			\toprule
			\multirow{3}{*}{Methods}   &		\multirow{3}{*}	{\begin{tabular}[c]{@{}c@{}} Attack \\ cost (ms) \end{tabular}}	&  \multicolumn{4}{c}{\centering SiamRPN~\cite{siamrpn}}  		
			

			& \multicolumn{4}{c}{SiamRPN++~\cite{siamrpnpp}} \\
			\cmidrule(l{2pt}r{16pt}){3-6} \cmidrule(l{2pt}r{16pt}){7-10} 
			& &  \multicolumn{2}{c}{\centering $\epsilon$ = 8}  
			& \multicolumn{2}{c}{\centering $\epsilon$ = 16} 
			&  \multicolumn{2}{c}{\centering $\epsilon$ = 8}  
			& \multicolumn{2}{c}{\centering $\epsilon$ = 16} \\ 
			\cmidrule(l{2pt}r{7pt}){3-4} \cmidrule(l{2pt}r{7pt}){5-6}  \cmidrule(l{2pt}r{7pt}){7-8}   \cmidrule(l{2pt}r{7pt}){9-10} 
			& & S  ($\uparrow$)			& P  ($\uparrow$)			& S($\uparrow$)			& P($\uparrow$)		& S  ($\uparrow$)			& P  ($\uparrow$)			& S($\uparrow$)			& P($\uparrow$)
			\\		
			\midrule
			Normal & 0  &65.1\%&85.5\%&65.1\% & 85.1\%  & 69\%  &  91\% &  69\% &  91\%\\	
			\midrule
			CSA-T &  3 & 54.3\% & 72.0\%&55.0\%& 73.6\%  &60.8\% & 80.9\% & 59.0\% & 79.3\%\\	
			CSA-S &   4720 & 34.5\% & 51.4\%&32.5\%&45.8\% & 34.8\% & 49.3\% & 37.1\% & 53.1\%\\	
			CSA-TS &  4720 & 30.7\% &46.2\%& 31.0\%&46.8\% &  33.1\% & 48.4\% & 34.7\% & 51.0\% \\	
			\midrule	 					
			$\textrm{Ours}_f$ & 8 & 46.3\%&66.5\%&  41.8\% & 58.8\% &  38.7\% & 54.0\% & 38.8\% & 54.9\%\\	
			$\textrm{Ours}$ &  8 & {\bf 20.8\%}& \bf{25.7\%}&\bf{14.5\%}& \bf{17\% } & \bf{14.9\%}& \bf{18.3\%}&\bf{7.0\%}& {\bf 8.0\%}\\	
			\bottomrule
		\end{tabular} 
	}
	\caption{{ \footnotesize Untargeted attack results on {\bf OTB100}~\cite{otb100}.}}
	\label{table:untargeted_otb}
\end{table}

\comment{
\begin{table}[t]
	\centering
			\resizebox{0.8\columnwidth}{!}{%
				\begin{tabular}{@{}c@{\hskip 0.3in}cc@{\hskip 0.3in}cc}
					\toprule
					\multirow{2}{*}{Networks}   
					&  \multicolumn{2}{c}{\centering $\epsilon$ = 8}  
					& \multicolumn{2}{c}{\centering $\epsilon$ = 16} 
				
					\\  \cmidrule(l{-5pt}r{16pt}){2-3} \cmidrule(l{-5pt}r{8pt}){4-5} 
 					& Success  ($\uparrow$)
 					& Precision  ($\uparrow$)
 					& Success($\uparrow$)
 					& Precision($\uparrow$)
 					\\
 					\midrule
 					Original & 1e-5 &0.65&0.08&100\% \\	
 					\midrule
 					 CSA-AT & 1e-5 &0.65&0.08&100\% \\	
 					 CSA-AS & 1e-5 &0.65&0.08&100\%\\	
 					 CSA-ATS & 1e-5 &0.65&0.08&100\%\\	
 					 					 				 					\midrule	 					
 					 					 					 					 					Ours & 1e-5 &0.65&0.08&100\%\\	
 					 					 					 					 					 					Ours & 1e-5 &0.65&0.08&100\%\\	
 	 				\bottomrule
				\end{tabular} %
			}
	\caption{  { \footnotesize Untargeted attack results on OTB100~\cite{otb100}}}
	\label{tbl:untargetedotb}
\end{table}

}

\noindent{\bf Results on VOT2018.} As shown in Table~\ref{tbl:untargeted_vot}, our approach is equally effective on VOT2018~\cite{vot2018}.

\begin{figure}[!h]
	\begin{center}
		\includegraphics[width=0.97\linewidth]{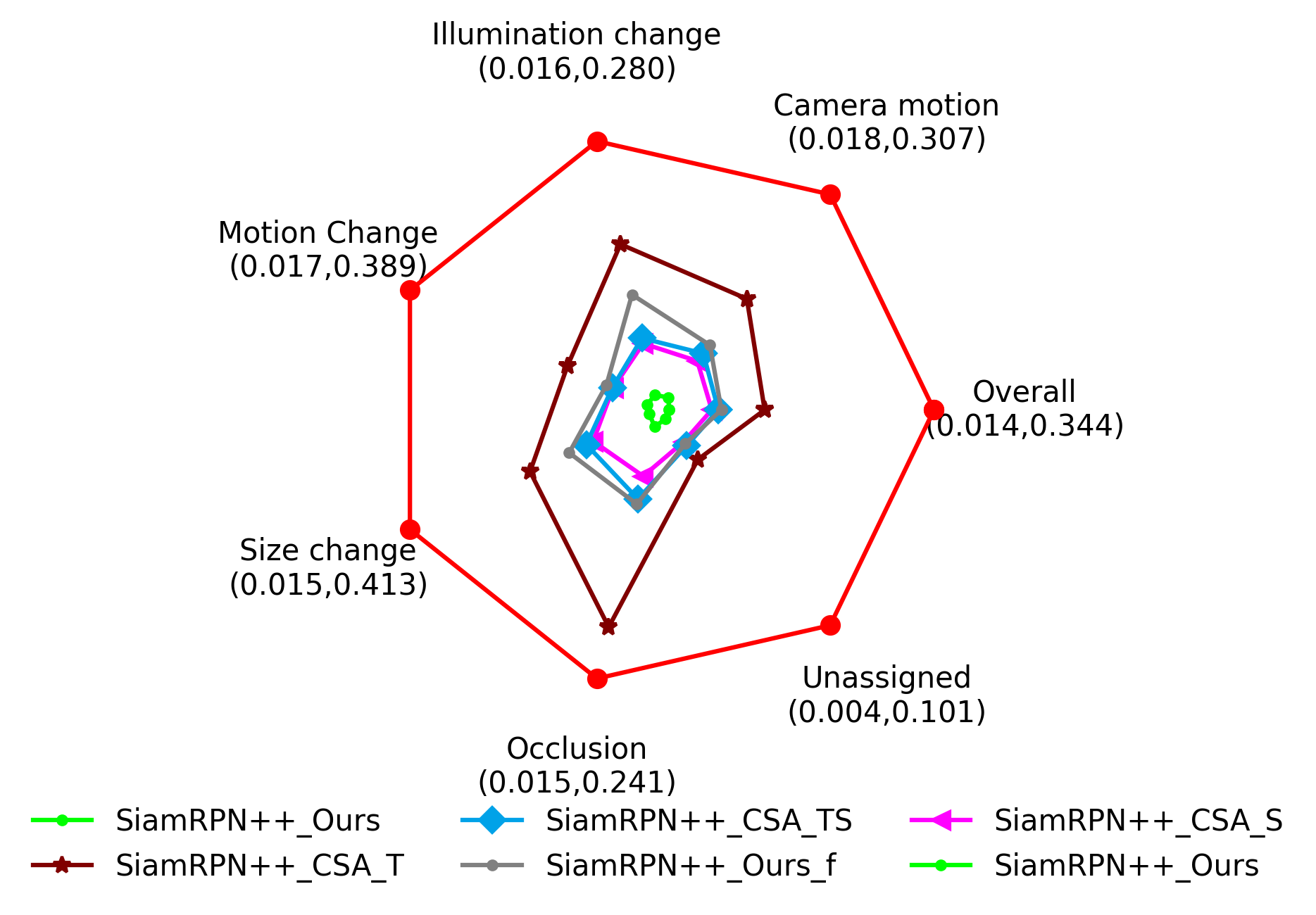}
	\end{center}
	\vspace{-3mm}
	\caption{\small{Quantitative analysis of the influence of different attributes on VOT2018~\cite{vot2018} with and without attacks.} \ms{The closer a point is to the center, the least successful the tracker is.}\MS{Correct?}\KN{Yes, closer to center implies more powerful attack}}
	\label{fig:attributevot}
\end{figure}

Note that, in this dataset, the accuracy does not significantly decrease because the tracker restarts as soon the object moves away from the ground-truth box. Importantly, unlike CSA, we compute the perturbation only from the template extracted from the first frame, even when the tracker restarts with a new template after a failure.  In other words, our perturbation computed from the initial template remains effective and transferable to temporally-distance search regions. 
As shown in Table~\ref{tbl:untargeted_vot}, for SiamRPN++~\cite{siamrpnpp} with $\epsilon=16$, our approach significantly decreases the EAO, which is the primary metric to rank trackers, from  0.261 to \comment{0.220} \kn{0.020} while increasing the number of restarts from 94 to 1249. In Figure~\ref{fig:attributevot}, we visually compare the results of our approach  and of the baselines in different conditions, defined as attribute values in the dataset. $\textrm{Ours}$  generalizes nicely to all these scenarios. 


\begin{table}[t]
	\centering
	\resizebox{1.0\columnwidth}{!}{
		\begin{tabular}{lc|ccccc|ccccc}
			\toprule
			&  	\multirow{2}{*}{Method}   &	\multirow{2}{*}	{\begin{tabular}[c]{@{}c@{}} Attack \\ cost (ms) \end{tabular}}	
			&   \multicolumn{4}{c|}{\centering $\epsilon$ = 8}  
			&	\multirow{2}{*}	{\begin{tabular}[c]{@{}c@{}} Attack \\ cost (ms) \end{tabular}}	
			& \multicolumn{4}{c}{\centering $\epsilon$ = 16}  \\ 
			\cmidrule(l{0pt}r{0pt}){4-7}  
			\cmidrule(l{2pt}r{7pt}){9-12}  
			& & & A  ($\uparrow$)		& R  ($\downarrow$)		& EAO($\uparrow$)	 & Re($\downarrow$)	&  & A($\uparrow$)		& R($\downarrow$)		& EAO($\uparrow$)  & Re($\downarrow$) \\		
			\midrule
			
			\multirow{6}{*}{\rotatebox{90}{SiamRPN++~\cite{siamrpnpp}}} &   
			
			Normal & 0  & 0.60  & 0.32  & 0.340 & 69  & 0 & 0.60 & 0.32 & 0.34 & 69  \\
			\cmidrule{2-12}  
			& CSA-T &     9 &  0.56 & 0.57 & 0.215 & 121 & 15 & 0.58 & 1.06 & 0.133 & 226   \\	
			& CSA-S & 2832 & 0.49 & 1.99 & 0.074 & 423 & 2832 &  \textbf{0.48} & 2.20 & 0.067 & 464\\	
			& CSA-TS & 2832 &  \textbf{0.46} & 1.70 & 0.086 & 363 & 2856 & 0.49 & 1.98 & 0.075 & 421\\
			\cmidrule{2-12}  
			& $\textrm{Ours}_f$ &  8 & 0.55 & 1.55 & 0.095 & 330 & 8 & 0.50 & 1.80 & 0.079 & 384\\	
			& $\textrm{Ours}$  & 8 & 0.55 &  \textbf{7.14} &  \textbf{0.017} &  \textbf{1524} & 8 & 0.52 &  \textbf{7.82} &  \textbf{0.014} &  \textbf{1669} \\			
			\midrule
			\multirow{6}{*}{\rotatebox{90}{SiamRPN~\cite{siamrpn}}} &   
			Normal &   0 & 0.57 & 0.44 & 0.261 & 94 & 0 & 0.57 & 0.44 & 0.261 & 94 \\
			\cmidrule{2-12}  
			& CSA-T &  15 & 0.56 & 1.04 & 0.132 & 222 &  15 & 0.56 & 1.10 & 0.126 & 234 \\	
			& CSA-S &  2832 & 0.45 & 1.76 & 0.080 & 376 & 2832& 0.46 & 1.95 & 0.077 & 417 \\	
			& CSA-TS &  2853 & \textbf{0.43} & 1.63 & 0.082 & 348 & 2856 &  \textbf{0.43} & 1.90 & 0.076 & 405\\
			\cmidrule{2-12}  
			& $\textrm{Ours}_f$ &  8 & 0.51 & 1.45 & 0.103 & 310 & 8 & 0.50 & 1.36  & 0.110 & 291 \\	
			
			& $\textrm{Ours}$ &  8 & 0.55 &  \textbf{5.06} &  \textbf{0.027} &  \textbf{1080} & 8 & 0.53 &  \textbf{5.85} &  \textbf{0.020} &  \textbf{1249} \\			
			
			
			\bottomrule
		\end{tabular} 
	}
	\caption{  { \small Untargeted attack results on {\bf VOT2018}~\cite{vot2018}.}\MS{Why are there no bold numbers?}\KN{Done}}
	\label{tbl:untargeted_vot}
\end{table}

\noindent{\bf Results on UAV123.} In Figure~\ref{fig:untargeted_UAV}, we provide success  and precision plots obtained by varying the overlap and location error thresholds, respecitvely, with $\epsilon=16$.   We observe a similar trend here: $\textrm{Ours}_f$ yields results that are comparable to those of CSA\_TS despite being much faster, and $\textrm{Ours}$ significantly outperforms all the baselines at all the thresholds. For example, at a 20 pixel threshold, while CSA\_TS yields a tracking precision of 0.504 precision, $\textrm{Ours}$ decreases it to 0.139.

\begin{table}[t]
	\centering
	\resizebox{0.96\columnwidth}{!}{
		\begin{tabular}{cccccccccc}
			\toprule
			\multirow{3}{*}{Networks}   &		\multirow{3}{*}{ Speed}  	&  \multicolumn{4}{c}{\centering SiamRPN~\cite{siamrpn}}  		
			& \multicolumn{4}{c}{SiamRPN++~\cite{siamrpnpp}} \\
			\cmidrule(l{2pt}r{16pt}){3-6} \cmidrule(l{2pt}r{16pt}){7-10} 
			& &  \multicolumn{2}{c}{\centering $\epsilon$ = 8}  
			& \multicolumn{2}{c}{\centering $\epsilon$ = 16} 
			&  \multicolumn{2}{c}{\centering $\epsilon$ = 8}  
			& \multicolumn{2}{c}{\centering $\epsilon$ = 16} \\ 
			\cmidrule(l{2pt}r{7pt}){3-4} \cmidrule(l{2pt}r{7pt}){5-6}  \cmidrule(l{2pt}r{7pt}){7-8}   \cmidrule(l{2pt}r{7pt}){9-10} 
			& & S  ($\uparrow$)			& P  ($\uparrow$)			& S($\uparrow$)			& P($\uparrow$)		& S  ($\uparrow$)			& P  ($\uparrow$)			& S($\uparrow$)			& P($\uparrow$)
			\\		
			\midrule
			Normal & 0  &65.1\%&85.5\%&65.1\% & 85.1\%  & 69\%  &  91\% &  69\% &  91\%\\	
			\midrule
			CSA-T &  XX & 54.3\% & 72.0\%&55.0\%& 73.6\%\\	
			CSA-S &   XX & 34.5\% & 51.4\%&32.5\%&45.8\%\\	
			CSA-TS &  XX & 30.7\% &46.2\%& 31.0\%&46.8\% \\	
			\midrule	 					
			Ours & 8 & 46.3\%&66.5\%&  41.8\% & 58.8\% &  38.7\% & 54.0\% & 38.8\% & 54.9\%\\	
			Ours+D &  8 & 20.8\%& 25.7\%&14.5\%& 17\%  & 14.9\%& 18.3\%&7.0\%&8.0\%\\	
			\bottomrule
		\end{tabular} 
	}
	\caption{  { \footnotesize Untargeted attack results on {\bf OTB100}~\cite{otb100}}}
	\label{tbl:untargeted_otb}
\end{table}

\comment{
\begin{table}[t]
	\centering
			\resizebox{0.8\columnwidth}{!}{%
				\begin{tabular}{@{}c@{\hskip 0.3in}cc@{\hskip 0.3in}cc}
					\toprule
					\multirow{2}{*}{Networks}   
					&  \multicolumn{2}{c}{\centering $\epsilon$ = 8}  
					& \multicolumn{2}{c}{\centering $\epsilon$ = 16} 
				
					\\  \cmidrule(l{-5pt}r{16pt}){2-3} \cmidrule(l{-5pt}r{8pt}){4-5} 
 					& Success  ($\uparrow$)
 					& Precision  ($\uparrow$)
 					& Success($\uparrow$)
 					& Precision($\uparrow$)
 					\\
 					\midrule
 					Original & 1e-5 &0.65&0.08&100\%\\	
 					\midrule
 					 CSA-AT & 1e-5 &0.65&0.08&100\%\\	
 					 CSA-AS & 1e-5 &0.65&0.08&100\%\\	
 					 CSA-ATS & 1e-5 &0.65&0.08&100\%\\	
 					 					 				 					\midrule	 					
 					 					 					 					 					Ours & 1e-5 &0.65&0.08&100\%\\	
 					 					 					 					 					 					Ours & 1e-5 &0.65&0.08&100\%\\	
 	 				\bottomrule
				\end{tabular} %
			}
	\caption{  { \footnotesize Untargeted attack results on OTB100~\cite{otb100}}}
	\label{tbl:untargetedotb}
\end{table}

}

\noindent{\bf Results on VOT2018-LT.} VOT2018-LT~\cite{vot2018} is dedicated to long-term tracking, and thus evaluating the strength of an attack should take into account not only the performance metrics but also the speed of the tracker. This is because, when the detection confidence is less than $0.8$ the tracker falls back to using a larger search region of $850 \times 850$ and thus becomes slower. This also means that the attack should not only  minimize the probability of the true target but also activate adversarial boxes with the confidence larger than $0.8$ so that search images are not resized.

As can be seen from Table~\ref{tbl:untargeted_votLT}, Ours yields F1 scores that are similar to those of CSA\_T, but lower frame-rates, showing that it tends to produce confidences that are lower than 0.8. By contrast, $\textrm{Ours}$ not only decreases the F1 score to significantly lower values than all the baselines, but it also maintains high frame-rates.



\begin{table}[t]
	\centering
	\resizebox{0.9\columnwidth}{!}{
		\begin{tabular}{cccccccccc}
			\toprule
			\multirow{3}{*}{Networks}   &			\multirow{3}{*}	{\begin{tabular}[c]{@{}c@{}} Attack \\ cost (ms) \end{tabular}}		&  \multicolumn{4}{c}{\centering DASiamRPN\_LT~\cite{siamrpn}}  		
			& \multicolumn{4}{c}{SiamRPN++\_LT~\cite{siamrpnpp}} \\
			\cmidrule(l{2pt}r{16pt}){3-6} \cmidrule(l{2pt}r{16pt}){7-10}

			& &  \multicolumn{2}{c}{\centering $\epsilon$ = 8}  
			& \multicolumn{2}{c}{\centering $\epsilon$ = 16} 
			&  \multicolumn{2}{c}{\centering $\epsilon$ = 8}  
			& \multicolumn{2}{c}{\centering $\epsilon$ = 16} \\ 
			\cmidrule(l{2pt}r{7pt}){3-4} \cmidrule(l{2pt}r{7pt}){5-6}    \cmidrule(l{2pt}r{7pt}){7-8} \cmidrule(l{2pt}r{7pt}){9-10}  
			& & 			    \multicolumn{1}{c}{F1($\uparrow$)}	 & \multicolumn{1}{c}{fps($\uparrow$)}			  &\multicolumn{1}{c}{F1($\uparrow$)}	 & \multicolumn{1}{c}{fps($\uparrow$)}& \multicolumn{1}{c}{F1($\uparrow$)}	 & \multicolumn{1}{c}{fps($\uparrow$)}			 & \multicolumn{1}{c}{F1($\uparrow$)}	 & \multicolumn{1}{c}{fps($\uparrow$)} \\
			
			\midrule
			
			Normal & 0 &  0.48  &  23.3 &0.48  & 23.3 & 0.61 & 16.3 & 0.61  & 16.3  \\
			\midrule
			CSA\_T  & 3 & 0.41  &  20.3 & 0.42 & 27.2  &  0.47 & 13.6 &  0.28 & 8.5 \\
			CSA\_S  & 33552 &  0.38 &27.2 & 0.37 & 20.3 &0.48 &  \textbf{6.4} &0.46    & 7.1\\
			CSA\_TS  & 33552 & 0.39 &   \textbf{16.3}&0.38 &  \textbf{16.5}& 0.47 & 8.1 & 0.46 & 7.4\\
									\midrule
			$\textrm{Ours}_f$  & 8 & 0.41 &20.3 & 0.42 & 20.4 &0.53 &9.0 &0.50 &  \textbf{6.6}\\
			$\textrm{Ours}$& 8 &  \textbf{0.35} & 20.3&  \textbf{0.26} & 20.5&  \textbf{0.18} &  12.4 & \textbf{0.08} & 16.3\\

			\bottomrule
		\end{tabular} 
\	}
	\caption{  { \small Untargeted attack results on {\bf VOT2018-LT}~\cite{vot2018}.}}
	\label{tbl:untargeted_votLT}
\end{table}


\comment{
\begin{table}[t]
	\centering
			\resizebox{0.8\columnwidth}{!}{%
				\begin{tabular}{@{}c@{\hskip 0.3in}cc@{\hskip 0.3in}cc}
					\toprule
					\multirow{2}{*}{Networks}   
					&  \multicolumn{2}{c}{\centering $\epsilon$ = 8}  
					& \multicolumn{2}{c}{\centering $\epsilon$ = 16} 
				
					\\  \cmidrule(l{-5pt}r{16pt}){2-3} \cmidrule(l{-5pt}r{8pt}){4-5} 
 					& Success  ($\uparrow$)
 					& Precision  ($\uparrow$)
 					& Success($\uparrow$)
 					& Precision($\uparrow$)
 					\\
 					\midrule
 					Original & 1e-5 &0.65&0.08&100\%\\	
 					\midrule
 					 CSA-AT & 1e-5 &0.65&0.08&100\%\\	
 					 CSA-AS & 1e-5 &0.65&0.08&100\%\\	
 					 CSA-ATS & 1e-5 &0.65&0.08&100\%\\	
 					 					 				 					\midrule	 					
 					 					 					 					 					Ours & 1e-5 &0.65&0.08&100\%\\	
 					 					 					 					 					 					Ours & 1e-5 &0.65&0.08&100\%\\	
 	 				\bottomrule
				\end{tabular} %
			}
	\caption{  { \footnotesize Untargeted attack results on OTB100~\cite{otb100}}}
	\label{tbl:untargetedotb}
\end{table}

}

\subsection{Targeted attacks}

Let us now turn to targeted attacks. In this context, we consider two scenarios:
\begin{enumerate}
\item The attacker forces the tracker to follow a fixed direction. We illustrate this with 4 different directions, consisting of shifting the box by ($\pm3$, $\pm3$) pixels in each consecutive frames, corresponding to the four directions $+45^\circ$, $-45^\circ$, $+135^\circ$, $-135^\circ$.
\item The attacker seeks for the tracker to follow a more complicated trajectory. To illustrate this, we force the tracker to follow the ground-truth trajectory with a fixed offset  $(\Delta_x, \Delta_y)$. We set this offset to ($\pm80$, $\pm80$) for the short-term tracking datasets and to ($\pm150$, $\pm150$) for VOT2018-LT to account for the larger resolution of its videos. 
\end{enumerate}
In both cases, we pre-compute perturbations corresponding $K=12$ diverse directions and use them to force the tracker to follow the target trajectory.  To this end, we first sample $K$ points at a distance $d=4$ from the true object center and, for each, synthesize a condition mask $\bM_i \in \{0,1\} ^ {(W \times H )}$ whose active region is centered at the sampled point. We then feed each such mask with the template to obtain directional perturbations, which we will then transfer to the search images. At each frame, we compute the direction the tracker should move in and use the precomputed perturbation that is closest to this direction.

We report the results for our two attack scenarios in Table~\ref{tbl:targeted} for all datasets with the SiamRPN++~\cite{siamrpnpp} tracker. Since the goal is to follow the desired trajectory, we report the precision for a location threshold of 20 pixels for all short-term datasets and 50 pixels for the long-term one. For direction-based targets,  
our perturbations allow us to effectively follow the target trajectory. For example, on OTB100~\cite{otb100}, our approach yields a precision of $53.7\%$ on average on the 4 directions, while the tracker has an original precision of $91\%$  on the original  data. 
Similarly, on VOT2018, our targeted attacks achieve a precision of $69.2\% $ on average, compared to the original tracker's precision of $72.8\%$.  
For offset-based targets,  which are more challenging than direction-based ones, our approach yields precision scores of $45.6\%$, $56.2\%$, $30.9\%$ and $24.9\%$ on OTB100, VOT2018, UAV, and VOT2018-LT, respectively. 

\begin{figure*}[h]
	\footnotesize
	\centering
      \includegraphics[width=17cm]{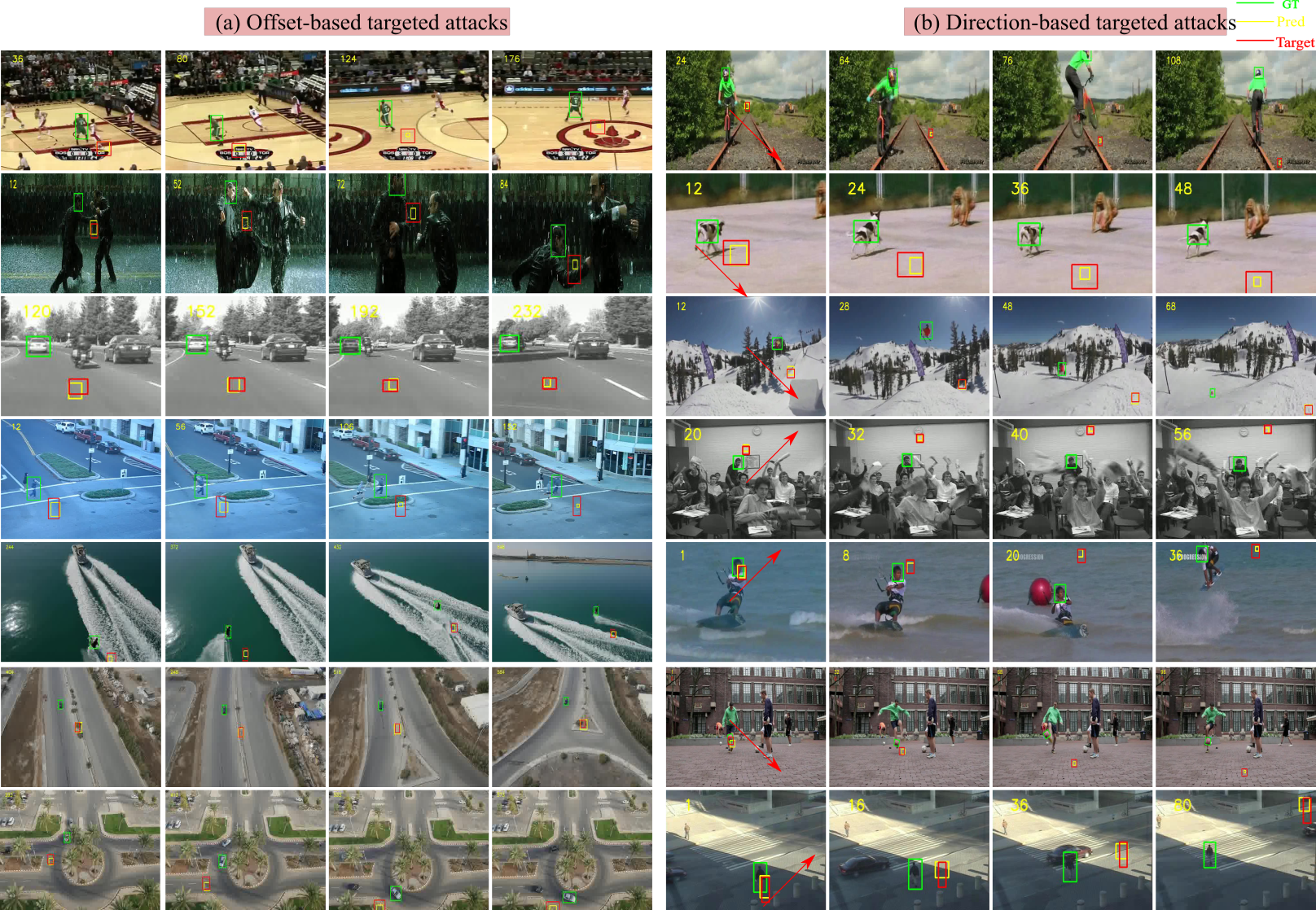}
	\caption{{\bf Qualitative results.} {\bf (a)} Results of targeted attacks where the tracker is forced to follow the ground-truth with a fixed offset. {\bf (b)} The tracker is forced to move in a constant direction, indicated by the red arrow.}
	\label{fig:targeted_results}
\end{figure*}

{Figure~\ref{fig:targeted_results} shows the results of targeted attacks on various datasets with SiamRPN++~\cite{siamrpnpp}. The results on the left, where the tracker follows the ground-truth trajectory with an offset, illustrate the real-world applicability of our attacks, where one could force the tracker to follow a realistic, yet erroneous path. Such realistic trajectories can deceive the system without raising any suspicion.}

\begin{figure}[t]
	\footnotesize
	\centering
      \includegraphics[width=\linewidth]{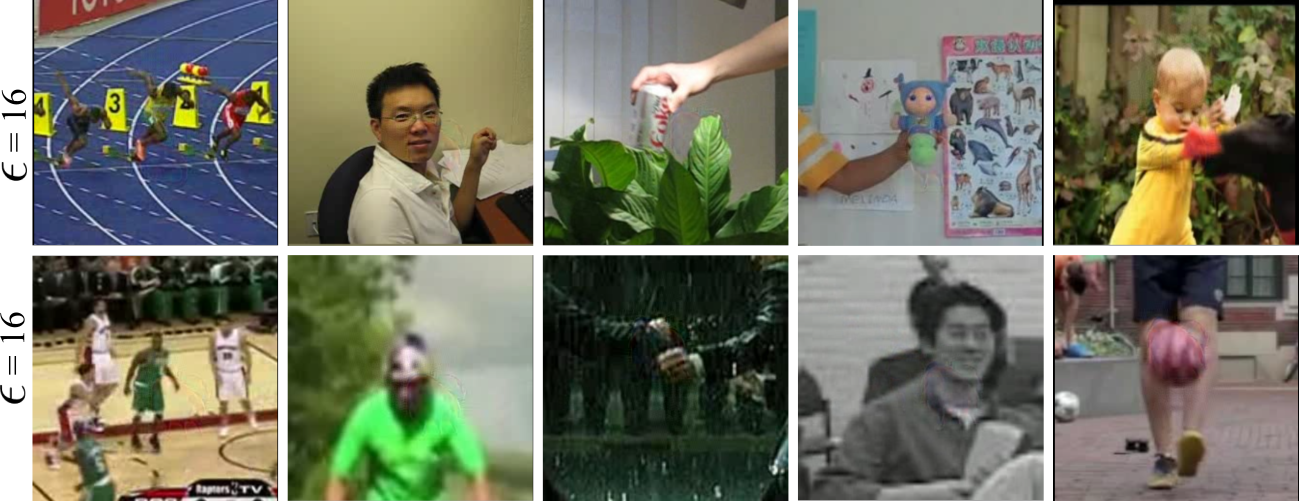}
	\caption{{\bf Perturbed search regions.}  We visualize adversarial search regions \kn{for targeted attacks} obtained with our framework for $\epsilon=16$ with SiamRPN++~\cite{siamrpnpp}.}
	\label{fig:perceptability}
\end{figure}

\ms{In Figure~\ref{fig:perceptability}, we visualize adversarial search regions obtained with different $l_{\infty}$ bound values for targeted attacks. We observe that, for  $\epsilon=16$, our results remain imperceptible thanks to our similarity loss.  We provide more qualitative results and perceptibility analyses in the supplementary material.}


\begin{table}[t]
	\centering
	\resizebox{0.98\columnwidth}{!}{
		\begin{tabular}{cccccccc}
			\toprule
			 & \multirow{1}{*}{Datasets}   &	\multirow{1}{*}	{\begin{tabular}[c]{@{}c@{}} Normal  \end{tabular}}	 
			 &   \multicolumn{1}{c}{\centering  $ +45^\circ$}  & \multicolumn{1}{c}{\centering  $-45^\circ$} 	&  \multicolumn{1}{c}{\centering  $+135^\circ$}  			& \multicolumn{1}{c}{\centering  $-135^\circ$}	  & Avg. 		\\ 
		
		\midrule
			
		\multirow{4}{*}{\begin{tabular}[c]{@{}c@{}} Direction-\\ based  \\ targets  \end{tabular}} &   	
			
			OTB100 ~\cite{otb100}   & 91\% & 57.7\% & 47.5\% & 55.0\% & 54.7\% & 53.7\%  \\ 
		&	VOT2018 ~\cite{vot2018}   & 72.8\%\% & 68.2\% & 71.0\% & 67.5\% & 69.9\%  &  69.2\%\\
		&	UAV123 ~\cite{uav123}   & 81.3\% & 43.1\% & 41.0\% & 49.2\% & 47.3\% & 45.2\% \\ 
		&	VOT2018-LT ~\cite{vot2018}   & 77.2\% & 35.6\% & 35.1\% & 35.7\% &35.0\% &  35.3\%\\ 
		
		\midrule
		
	& \multirow{1}{*}{Datasets}   &	\multirow{1}{*}	{\begin{tabular}[c]{@{}c@{}} Normal  \end{tabular}}	 	& \multicolumn{1}{c}{\centering  ($\Delta_x$, $\Delta_y$)} 			&  \multicolumn{1}{c}{\centering ($\Delta_x$, -$\Delta_y$)}  			& \multicolumn{1}{c}{\centering  (-$\Delta_x$, +$\Delta_y$)}			& \multicolumn{1}{c}{\centering  (-$\Delta_x$, -$\Delta_y$)} & Avg.\\
	\midrule

		\multirow{4}{*}{\begin{tabular}[c]{@{}c@{}} Offset-\\based\\ targets \end{tabular}} &   	
		OTB100 ~\cite{otb100}   & 91\% &  50.7\% & 45.4\% & 44.2\% & 42.0\% & 45.6\% \\
		&VOT2018 ~\cite{vot2018}   & 72.8\% &  55.2\% & 53.9\% & 58.2\% & 57.5\%  & 56.2\%\\
		&	UAV123 ~\cite{uav123}   & 81.3\% & 36.2\% & 38.4\% & 44.1\% & 40.9\% & 39.9\%\\
		&VOT2018-LT ~\cite{vot2018}   & 77.2\%  &  25.7\% & 25.1\% & 28.5\% & 21.1\% & 25.1\%\\

			\bottomrule
		\end{tabular} 
	}
	\caption{{ \footnotesize {\bf  Targeted attack results.} We report precision scores with $\epsilon=16$.   $\Delta_x\in\{80,150\}$ and $\Delta_y\in\{80,150\}$ depending on the dataset.}}
	\label{tbl:targeted}
\end{table}

\MS{Is this for targeted or untargeted? I suggest you move this to the corresponding subsection. In fact, the paragraph above is also showing qualitative results, so I don't see the point of having a separate sub-section here.}
\KN{merged into above subsection}

\subsection{Ablation Study}

In this section, we analyze the impact of each loss term of our framework.  To do so, we use the SiamRPN++ tracker~\cite{siamrpnpp} and perform untargeted attacks. In Table~\ref{tbl:ablation}, we report our results on OTB100~\cite{otb100} and VOT2018~\cite{vot2018} with different combination of loss terms, where $\mathcal{L}_{fool}^{cls}$ and $\mathcal{L}_{fool}^{reg}$ represent the classification and regression components of the fooling loss of Eq.~\ref{eq:foolingloss},  $\mathcal{L}_{shift}^{cls}$ and $\mathcal{L}_{shift}^{reg}$  represent the same terms for the shift loss of Eq.~\ref{eq:shiftloss}.  To summarize,  
while all loss terms are beneficial, the classification-based terms are more effective than regression-based ones.  For example, using either $\mathcal{L}_{fool}^{cls}$ or  $\mathcal{L}_{fool}^{cls}$  has more impact than $\mathcal{L}_{fool}^{reg}$ or $\mathcal{L}_{fool}^{reg}$.  {In Table~\ref{tbl:ablation_shift}, we study the impact of the shift distance $d$ in Eq.~\ref{eq:shiftloss} on the performance of untargeted attacks. For a feature map of size $25 \times 25$ for SiamRPN++~\cite{siamrpnpp}, the performance of our approach is stable for a drift in the range 4 to 8. However, for $d=2$, our attacks have less effect on the tracker, and for $d=10$, the influence of the attack decreases because of  the Gaussian prior used by the tracker.}

\begin{table}[t]
\centering

 \resizebox{0.8\columnwidth}{!}{%
 	\begin{tabular}{ccccccc}
    \toprule
     	\multirow{2}{*}{  $\mathcal{L}_{fool}^{cls}$ }&  	\multirow{2}{*}{$\mathcal{L}_{fool}^{reg}$}  & 	\multirow{2}{*}{$\mathcal{L}_{shift}^{cls}$ }  &	\multirow{2}{*}{$\mathcal{L}_{shift}^{reg}$ }    &   
     	
      \multicolumn{2}{c}{	OTB100~\cite{otb100} }  & VOT2018~\cite{vot2018} \\
     		\cmidrule(l{2pt}r{7pt}){5-6} \cmidrule(l{2pt}r{7pt}){7-7} 
     	
  & & & &  	 Success($\boldsymbol{ \uparrow}$)   & Precision($\boldsymbol{\uparrow}$)  & EAO($\boldsymbol{\uparrow}$) \\ 
     \midrule

     -&-&-&-& 69.5\% & 90.5\%  & 0.340 \\
     \ding{51}& -& -& -&  58.8\% & 66.5\% & 0.088\\			
     -& \ding{51}&-&-& 50.6\%& 72.6\% & 0.128\\			   
       -&-& \ding{51}&-& 50.1\%& 65.8\%  & 0.094 \\	
          -    &-&-& \ding{51}& 65.4\%&86.3\% & 0.024 \\			
                    \ding{51}  & \ding{51}&-&-& 38.7\%& 54.0\% & 0.095 \\			
            -          &-&  \ding{51}  & \ding{51}  & 40.5\% & 52.7\% & 0.063\\			
                        \ding{51} & - & \ding{51} &  -&  19.1\% &23.8\% & 0.024\\			
                     - &  \ding{51} & - & \ding{51}     & 46.4\%& 63.1\% & 0.092\\			
                                            \ding{51} &  \ding{51} &   \ding{51}& \ding{51}  &  \textbf{14.9\%} & \textbf{18.3\%} & \textbf{ 0.017}\\			
   \bottomrule
  \end{tabular}%
}
  \caption{\small{\textbf{Ablation study}. Contribution of each  loss for untargeted attacks  with SiamRPN++~\cite{siamrpnpp}. We set $\epsilon=8$.}}
  \label{tbl:ablation}
\end{table}

\comment{

\begin{table}[t]
	\centering
	
	\resizebox{0.8\columnwidth}{!}{%
		\begin{tabular}{cccccc}
			\toprule
			$\mathcal{L}_{fool}^{cls}$  & $\mathcal{L}_{fool}^{reg}$  & $\mathcal{L}_{shift}^{cls}$   &$\mathcal{L}_{shift}^{reg}$     & Success($\boldsymbol{ \uparrow}$)   & Precision($\boldsymbol{\uparrow}$)  \\ 
			\midrule

			-&-&-&-& 69.5\% & 90.5\% \\
			\ding{51}& -& -& -&  58.8\% & 66.5\%\\			
			-& \ding{51}&-&-& 50.6\%& 72.6\% \\			   
			-&-& \ding{51}&-& 50.1\%& 65.8\%  \\	
			-    &-&-& \ding{51}& 65.4\%&86.3\% \\			
			\ding{51}  & \ding{51}&-&-& 38.7\%& 54.0\% \\			
			-          &-&  \ding{51}  & \ding{51}  & 40.5\% & 52.7\%\\			
			\ding{51} & - & \ding{51} &  -&  19.1\% &23.8\%\\			
			- &  \ding{51} & - & \ding{51}     & 46.4\%& 63.1\%\\			
			\ding{51} &  \ding{51} &   \ding{51}& \ding{51}  & 14.9\% &18.3\%\\			
			\bottomrule
		\end{tabular}%
	}
	\caption{\small{\textbf{Ablation study}. Contribution of each  loss for untargeted atatcks on OTB100~\cite{otb100}  with SiamRPN++~\cite{siamrpnpp}. We set $\epsilon=8$.}}
	\label{tbl:ablation}
\end{table}

}

\begin{table}[t]
\centering

 \resizebox{0.8\columnwidth}{!}{%
 	\begin{tabular}{cccccccc}
    \toprule
        \multirow{2}{*}{Shift $d$ }   &   
     	
      \multicolumn{2}{c}{	OTB100~\cite{otb100} }  &    \multicolumn{4}{c}{VOT2018~\cite{vot2018}} \\
     		\cmidrule(l{2pt}r{7pt}){2-3} \cmidrule(l{2pt}r{7pt}){4-7} 
     	
  &   	 Success($\boldsymbol{ \uparrow}$)   & Precision($\boldsymbol{\uparrow}$)  &  A($\boldsymbol{\uparrow}$)  & R($\boldsymbol{\downarrow}$) & EAO($\boldsymbol{\uparrow}$) & Re($\boldsymbol{\downarrow}$)  \\ 
     \midrule   
     0 & 38.7\% & 54.0\%  & 0.549 & 1.54 & 0.095 & 330  \\
      2 &  16.6\%  & 20.5\%  &  0.558 & 5.25 & 0.026  & 1180\\
  	4 &  {14.9\% } &  18.3\% &  \textbf{0.552} & 7.31 &  \textbf{0.017} & 1524\\
     	6 & \textbf{13.8\% }&  \textbf{17.5\%}  & 0.564 & 7.27 &  \textbf{0.017} & 1552\\
     	8& 15.7\% & 19.6\% & 0.615 &  \textbf{7.32} & 0.018 &  \textbf{1563}\\
     	10 & 22.6\% & 28.7\% & 0.604 & 5.57 & 0.027 & 1189\\
   \bottomrule
  \end{tabular}%
}
  \caption{\small{\textbf{Ablation study}.  Effect of  $d$  in $\mathcal{L}_{shift}$ for untargeted attacks  with SiamRPN++~\cite{siamrpnpp}. We set $\epsilon=8$.}}
  \label{tbl:ablation_shift}
\end{table}

\subsection{Transferability Analysis}\label{sec:transferability}
To evaluate the transferability of our attacks, we apply the perturbation-generator trained on SiamRPN++~\cite{siamrpnpp} with a ResNet~\cite{resnet} backbone to three other state-of-the-art trackers: AlexNet-based SiamFC~\cite{siamfc}, ResNet-based SiamMask~\cite{siammask} and a similar  SiamRPN++(M)~\cite{siamrpnpp} but with a MobileNet backbone.  As can be seen in Table~\ref{tbl:transferattacks},  our approach transfers better than CSA\_T to SiamMask and SiamRPN++(M).

However, neither CSA nor our approach generalize well to SiamFC, which we conjecture to be due to SiamFC using a different backbone from that of SiamRPN++~\cite{siamrpnpp} and to its non-RPN-based architecture. We therefore believe that generating black-box attacks that generalize to arbitrary architectures would be an interesting avenue for future research.


\begin{table}[h]
	\centering
	\resizebox{0.96\columnwidth}{!}{
		\begin{tabular}{ccccccccccc}
			\toprule
			\multirow{2}{*}{Tracker}   &		\multirow{2}{*}{ Method}  	&  \multicolumn{4}{c}{\centering VOT2018~\cite{vot2018}}  		& \multicolumn{2}{c}{OTB100++~\cite{otb100}} \\
			\cmidrule(l{2pt}r{0pt}){3-6} \cmidrule(l{2pt}r{7pt}){7-8} 
			& &  A($\uparrow$) & R($\downarrow$)  & EAO($\uparrow$)& Re($\downarrow$)& S  ($\uparrow$)			& P  ($\uparrow$)		\\
			\midrule
			
				\multirow{5}{*}{SiamMask~\cite{siammask}} & Normal & 0.59 & 0.25 & 0.407 & 53 & 64.7\% &84.0\% \\
				
				& CSA\_T   &   0.53 & 0.36 &0.290 & 77 &  57.3\% &  78.6\%  \\
				& CSA\_S   & 0.35 & 2.81  & 0.045 & 601 &  21.5\%  & 36.4\% \\
				& CSA\_TS & \textbf{0.34} & 2.87 &  0.046 & 612  & 20.2\% & 34.6\%\\
		& Ours & 0.46 & 2.46  & 0.056 & 516&29.4\% & 44.4\%\\
	& $\textrm{Ours}_{shift}$&  {0.57} &  \textbf{6.08}  &  \textbf{0.023} &  \textbf{1298}& \textbf{16.0\%} & \textbf{ 20.6\%}\\

		\midrule
	\multirow{5}{*}{SiamRPN++(M)~\cite{siamrpnpp}} & Normal &  0.59 & 0.24 & 0.398 & 51  & 65.1\% & 85.5\%\\
	& CSA\_T & 0..56 & 0.45  & 0.265 & 95 & 61.5\% & 83.6\% \\
	& CSA\_S & 0.41 & 2.20 & 0.066 & 471 & 28.1\% &44.1\%\\
	& CSA\_TS  & \textbf{0.40} & 2.17 & 0.067 & 465& 26.3\% &42.5\%\\
		& Ours & 0.48 & 1.72  & 0.085 & 367&34.7\% & 52.7\%\\
	& $\textrm{Ours}_{shift}$ &0.52 & \textbf{5.15} & \textbf{0.028} & \textbf{1100}&\textbf{21.7\%} &\textbf{ 28.1\%}\\
		\midrule
	
	\multirow{5}{*}{SiamFC~\cite{siamfc}} & Normal &  0.49 & 0.55 & 0.211 & 118 & 56.9\% &76.2\%\\
	& CSA\_T   & 0.48 & 0.52 & 0.213 & 112 & 56.8\%& 75.8\%  \\
	& CSA\_S  & 0.49 & 0.56 & 0.210 & 120  & 57.2\%& 76.3\%\\
	& CSA\_TS & 0.49 &\textbf{0.57} & 0.211 & 121 & 57.5\%& 76.9\% \\
	& Ours & \textbf{0.48} & 0.55  & 0.201 & 118&\textbf{57.7\% }& 77.2\%\\
	& $\textrm{Ours}_{shift}$ &0.49 & 0.59 & \textbf{0.200} & \textbf{126}&55.9\% &\textbf{ 74.7\%}\\
	
	\bottomrule
		\end{tabular} 
	}
	\caption{ { \footnotesize {\bf Transferability analysis} on OTB100~\cite{otb100} with a generator trained on SiamRPN++~\cite{siamrpnpp}}. We set $\epsilon=8$.}
	\label{tbl:transferattacks}
\end{table}


\section{Conclusion}
We have presented a one-shot temporally-transferable perturbation framework to efficiently attack a VOT algorithm using only template information.  Our generates a single perturbation from the object template in only 8ms, which can then be transferred to attack the subsequent frames of the video. Furthermore, we have demonstrated that  conditioning the generator allows us to steer the tracker to follow any specified trajectory by precomputing a few diverse directional perturbations. We believe that our work highlights the vulnerability of object trackers and will motivate the researchers to design robust defense mechanisms in the future.


{\small
\bibliographystyle{ieee_fullname}
\bibliography{egbib}
}

\end{document}